\documentclass[9pt, sigconf]{acmart}
\usepackage{algorithmicx}
\usepackage{algorithm}
\usepackage{amsmath}
\usepackage{enumitem}
\usepackage{algpseudocode}
\usepackage{multirow}
\usepackage{epsfig}
\usepackage{multicol,lipsum,xparse}
\usepackage{subcaption}
\usepackage{amsmath}
\usepackage{stmaryrd}
\usepackage{color}
\usepackage{todonotes}
\usepackage[utf8]{inputenc}
\usepackage{nopageno}
\usepackage{graphicx}
\usepackage{tikz}
\title{Painting on Placement: Forecasting Routing Congestion using Conditional Generative Adversarial Nets}

\author{Cunxi Yu, Zhiru Zhang}
\affiliation{%
  \institution{CSL, Cornell University}
}
\email{cunxi.yu,zhiruz@cornell.edu}

\begin{document}

\newcommand{\pgftextcircled}[1]{
    \setbox0=\hbox{#1}%
    \dimen0\wd0%
    \divide\dimen0 by 2%
    \begin{tikzpicture}[baseline=(a.base)]%
        \useasboundingbox (-\the\dimen0,0pt) rectangle (\the\dimen0,1pt);
        \node[circle,draw,outer sep=0pt,inner sep=0.1ex] (a) {#1};
    \end{tikzpicture}
}

\copyrightyear{2019} 
\acmYear{2019} 
\setcopyright{acmcopyright}
\acmConference[DAC '19]{The 56th Annual Design Automation Conference 2019}{June 2--6, 2019}{Las Vegas, NV, USA}
\acmBooktitle{The 56th Annual Design Automation Conference 2019 (DAC '19), June 2--6, 2019, Las Vegas, NV, USA}
\acmPrice{15.00}
\acmDOI{10.1145/3316781.3317876}
\acmISBN{978-1-4503-6725-7/19/06}

\begin{abstract}

Physical design process commonly consumes hours to days for large designs, and routing is known as the most critical step. Demands for accurate routing quality prediction raise to a new level to accelerate hardware innovation with advanced technology nodes. This work presents an approach that forecasts the density of all routing channels over the entire floorplan, with features collected up to placement, using conditional GANs. Specifically, forecasting the routing congestion is constructed as an image translation (colorization) problem. The proposed approach is applied to a) placement exploration for minimum congestion, b) constrained placement exploration and c) forecasting congestion in real-time during incremental placement, using eight designs targeting a fixed FPGA architecture.

\end{abstract}
\maketitle

\section{Introduction}

As technology continues scaling, the complexity of physical design rules that are a series of parameters provided by manufacturers has been significantly increased. Physical design, the most runtime-critical design stage of Electronic Design Automation (EDA) flow, becomes more challenging with advanced technology nodes. Modern design closure process mostly requires many design iterations through full placement \& route (PnR) process, which is evidently expensive for large designs. Due to the long runtime and the lack of predictability of the physical design process, the challenges of design closure within short time-to-market raise to a new level. To overcome such barriers, predictive flow-level modeling and fast and accurate prediction techniques have very high value.

Recent years have seen an increasing employment of machine learning (ML) that target both front-end \cite{ziegler2017ibm,dai2018fccm,cunxi2018CNN,ustun2019fccm} and back-end \cite{xu2018subresolution,pui2017clock,ding2012epic,yu2015machine,xie2018routenet} design tools. For example, Xu \cite{xu2018subresolution} proposed a supervised learning based sub-resolution assist feature (SRAF) generator that is used to improve yield in the manufacturing process. Hotspot detection using has been studied using SVM-Kernels \cite{yu2015machine} and deep learning \cite{yang2018layout}. 
Specifically, for improving the quality of routing estimation at early stages, the most recent works mainly focus on a) forecasting routing congestion map \cite{pui2017clock} and b) routability prediction \cite{xie2018routenet,chan2017routability}. A machine learning based routing congestion prediction model is used for FPGA PnR \cite{pui2017clock}. However, this approach forecast the heat map by predicting the congestion only based on SLICEs. 
RouteNet \cite{xie2018routenet} predicts the number of Design Rule Violations (DRV) using transfer learning with ResNet18 as the pre-trained model. RoutNet also forecasts the locations of hotspots using a fully convolutional network (FCN). However, both works {xie2018routenet}\cite{chan2017routability} require the features collected at the routing stage.

This paper presents a novel approach that estimates the detailed routing congestion with a given placement solution for FPGA PnR. The proposed approach fully forecasts the routing congestion heat map using a conditional Generative Adversarial Nets (cGANs) model. The problem is constructed as \textit{image translation (colorization)}, where the {input} are the post-placement image, and the output is the congestion heat map obtained after detailed routing. The main contributions include \textbf{a)} Unlike the existing works that require features at routing stages, the proposed approach only requires features collected up to placement. \textbf{b)} This approach estimates the utilization of all routing channels by forecasting the full congestion heat map, instead of hotspots only. \textbf{c)} The analysis of training with L1 loss and skip connections cGANs are included in Section \ref{sec:l1_skip}. \textbf{d)} The proposed approach is applied to constrained placement exploration and real-time routing forecast while the design is being placed.
To the best of our knowledge, this is the first approach that forecasts the routing utilization (density) of all routing channels. This is also the first approach that estimates detailed routing congestion without any routing results.

\vspace{-3mm}
\section{Background}

\subsection{CNNs and FCNs}

Convolutional neural network is a class of deep artificial neural networks, which has been widely used in image classification \cite{krizhevsky2012imagenet}, language processing \cite{kim2014convolutional}, decision making \cite{silver2016mastering}, etc. The hidden layers of a CNN typically consist of convolutional layers, pooling layers, fully connected layers. Convolutional layers compute the local regions of the input and connected to local regions in the input, pooling layers perform downsampling over the spatial blocks, and the fully connected layer will finally compute the class scores that are used to produce the labels during inference. 

In contrast to CNNs, fully convolutional networks (FCNs) are built only with locally connected layers, which was proposed for semantic segmentation \cite{long2015fully}, without using any dense and pooling layer. FCN consists of downsampling path and upsampling path, where downsampling path captures semantic information and upsampling path recover the spatial information. To better upsample the spatial information produced by the downsampling layers, skip connections, i.e., bypass-connections that concatenate one layer in the downsampling path and one layer in the upsampling path, are used for transferring the local information cross different layers \cite{ronneberger2015u}. In this work, our deep neural network model leverages both CNNs and FCNs. The details of the model and discussions of skip connections are included in Section \ref{sec:approach}.

\begin{figure}[!htb]
\centering
\includegraphics[width=0.35\textwidth]{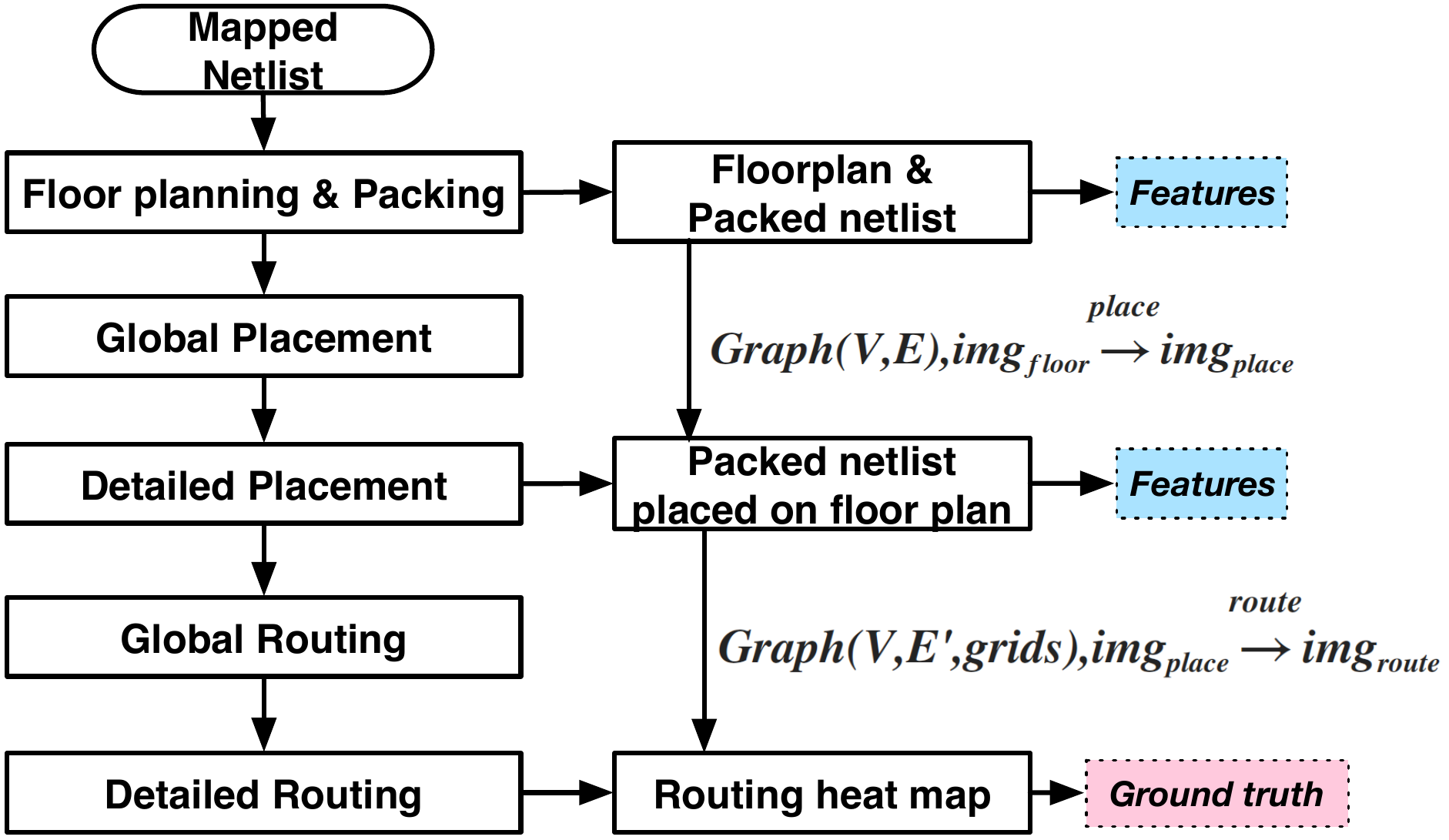}
\caption{Physical design flow and the concept of forecasting routing utilization using image translation.}
\vspace{-3mm}
\label{fig:pd_flow}
\end{figure}

\begin{figure*}[!htb]
\centering
\begin{minipage}{0.18\textwidth}
  \centering
\includegraphics[width=1\textwidth]{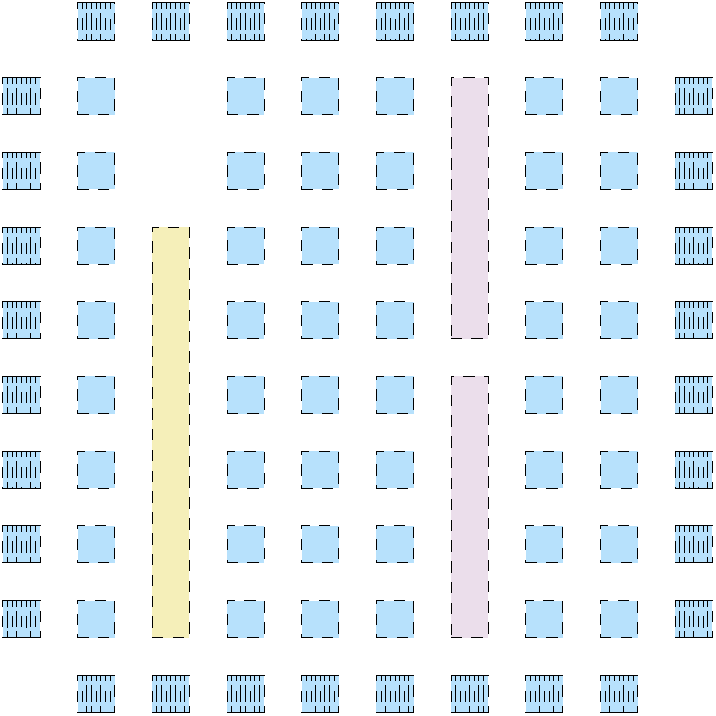}
\subcaption{$img_{floor}$}\label{fig:example1}
\end{minipage}
\hspace{0.5mm}
\begin{minipage}{0.18\textwidth}
  \centering
\includegraphics[width=1\textwidth]{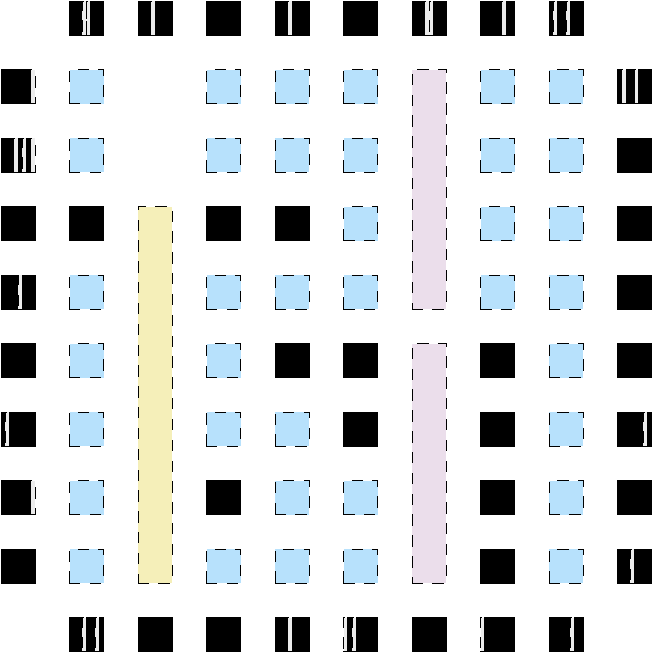}
\subcaption{$img_{place}$}\label{fig:example2}
\end{minipage}
\hspace{0.5mm}
\begin{minipage}{0.18\textwidth}
  \centering
\includegraphics[width=1\textwidth]{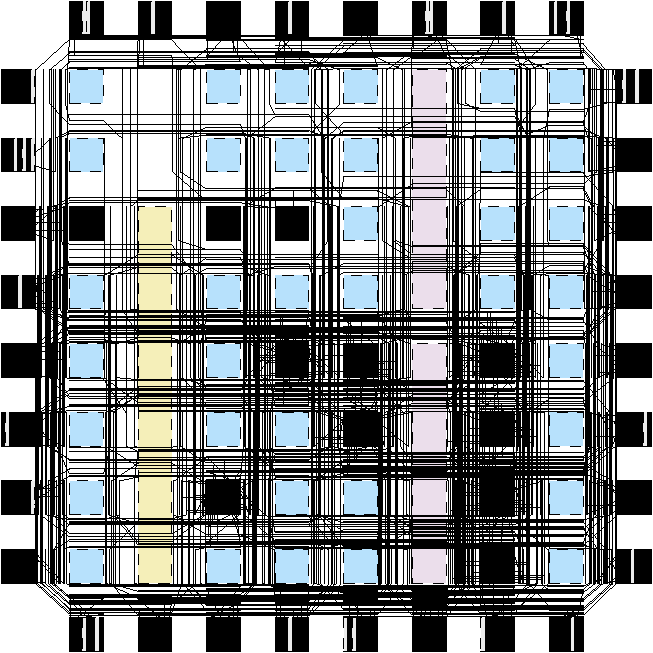}
\subcaption{Routing result}\label{fig:example3}
\end{minipage}
\hspace{0.5mm}
\begin{minipage}{0.19\textwidth}
  \centering
\includegraphics[width=1\textwidth]{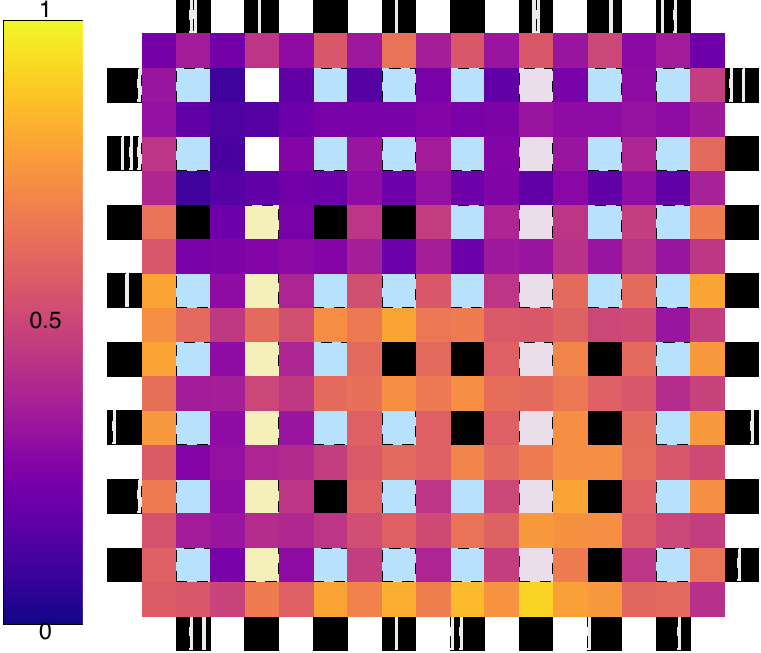}
\subcaption{$img_{route}$}\label{fig:example4}
\end{minipage}
\begin{minipage}{0.17\textwidth}
  \centering
\includegraphics[width=1\textwidth]{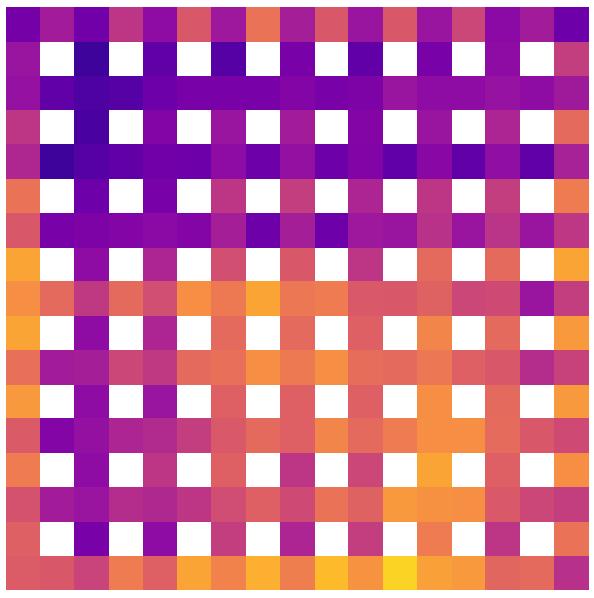}
\subcaption{$img_{route}$ - $img_{place}$}\label{fig:example5}
\end{minipage}
\caption{Motivating example of forecasting routing heat map as image colorization. a) floor plan image $img_{floor}$; b) post-placement image $img_{place}$; c) routing result; d) routing heat map image $img_{route}$ (ground truth); e) exact difference between $img_{place}$ and $img_{route}$. }
\label{fig:example}
\end{figure*}

\subsection{Physical design}\label{sec:pd}

Physical design is the process of transforming a circuit description into the physical layout, which describes the locations of the cells and the routs of the interconnections of the elements with respect to a floor plan. It includes design, verification, and validation at the layout level, and is known to be the most time-consuming process in the modern electronic design flow. In particular, routing is the slowest PnR stage and becomes more unpredictable as the technology advances \cite{yu2018retiming}. Hence, developing an accurate congestion prediction technique becomes critical.

One of the inputs for physical design process is a technology-mapped netlist, which is represented using directed graphs such that the cells are nodes $V$ and interconnects are edges $E$ (Figure \ref{fig:pd_flow}). Specifically for FPGA placement, it is a packed netlist where each cluster-based logic block (CLB) could contain one or more basic logic elements (BLEs). In Figure \ref{fig:pd_flow}, $Graph(V,E)$ refers to the packed netlist. Floorplanning is the process that allocates space for placement and routing by identifying the structures of the input netlist in order to meet the required performance and design rules. All the elements in $V$ are then placed within the floor plan. After placement, the nodes and edges in the graphs have a specific 2-D location on the floor plan, denoted as $Graph(V,E',grids)$, where $grids$ represent the 2-D locations of $V$. Meanwhile, the edges are updated with locations $E \rightarrow E'$. Finally, routing connects all the elements with respect to $Graph(V,E',grids)$. 

The intermediate results, i.e., floor planning, post-placement and post-routing results, can be visualized as images $\in \mathbb{R}^{w\times w \times 3}$, denoted as $img_{floor}$, $img_{place}$ and $img_{route}$, respectively. An important observation is that these images are incrementally changed while PnR proceeds: $Graph(V,E),img_{floor} \xrightarrow{} img_{place}$
or from post-placement to post-routing: $Graph(V,E',grids),img_{place} \xrightarrow{} img_{route}$
Based on this observation, the problem of forecasting routing heat map can be formulated as an \textit{image to image translation} problem. Specifically, the proposed approach generates the estimated routing heat map $img_{route}$ from $img_{place}$. To this end, we present a conditional generative adversarial networks (cGANs) based approach (Section \ref{sec:approach}) such that the generator learns a differentiable function $G$ such that maps $Graph(V,E),img_{place} \rightarrow img_{route}$.


\section{Routing forecast by "painting" placement}\label{sec:motivation}

We illustrate the concept of forecasting routing congestion as image translation using an example shown in Figure \ref{fig:example}. These images are generated by modifying VTR 8.0 \cite{luu2014vtr}. Figure \ref{fig:example1} shows the floor plan $img_{floor}$. There are three types of elements in $img_{floor}$: \textbf{a)} \textit{I/O pads}. The elements on each of the four sides of the floor plan, which are used for placing the inputs and outputs. For this specific FPGA architecture, each element includes eight ports that each of them can be used to place one input/output pad. \textbf{b)} \textit{CLB spots}. The six columns (1,3,4,5,7,8 columns) of elements surrounded by the input/output pads, which are used to place CLBs, i.e., $V$ in $Graph(V,E',grids)$. \textbf{c)} \textit{Memory and multiplier blocks}. The yellow element in the third column indicates the memory block and the pink bars in the seventh column indicate the multiplier block. Note that there could be more types of elements shown in the floor plan image for other FPGA architectures.  

Figure \ref{fig:example2} represents the post-placement result $img_{place}$. Compared to $img_{floor}$, the image has been updated by changing the pixels where CLBs and I/O pads are placed. Specifically, the corresponding pixels are filled with black pixels and the rest of the image remains the same. For example, in the second column, there is one CLB placed in the third row. The I/O pads may not be fully filled with black pixels since each of them contains eight ports. 
Similarly, the routing result $img_{route}$ can be represented on top of $img_{place}$ (Figure \ref{fig:example3}). Figure \ref{fig:example4} shows the congestion heat map which is used to visualize routing congestion by measuring the utilization of the routing channels. Compared to $img_{place}$, $img_{route}$ is updated by colorizing the routing channels pixels only, with respect to the utilization color bar. The pixel-to-pixel differences between $img_{route}$ and $img_{place}$ are shown in Figure \ref{fig:example5}.

One of the conditions required for a high-quality image to image translation is that the input and output images should have the same underlying structure, and mostly differ in the surface appearance \cite{pix2pix2016}. In other words, this requires that the structure of the input should be well aligned with the structure of the output. In this work, $img_{place}$ is the input image (other input features will be introduced in next section), and $img_{route}$ is the output image. The underlying image structures of these two images are almost identical. This offers the main motivation for leveraging image-to-image translation model for routing forecast.

\section{Approach}\label{sec:approach}

\subsection{GANs and cGANs}

Generative adversarial networks (GANs) are neural network models that are used in unsupervised machine learning tasks. GANs learn a transformation from random noise vector $z$ to a corresponding mapping $g$, denoted as $G(z)$, which implements a differentiable function that maps $z \rightarrow y$\cite{goodfellow2014generative}. GANs include two multilayer perceptrons, namely generator $G$ and discriminator $D$. The goal of discriminator $D$ is to distinguish between samples generated from the generator and samples from the training dataset. The goal of generator $G$ is to generate a mapping of input that cannot be distinguished to be true or false by the discriminator $D$. The network is trained in two parts and the loss function $L_{(G,D)}$ is shown in Equation \ref{eq:loss_gan}.

\begin{itemize}
\item train $D$ to maximize the probability of assigning the correct label to both training examples and samples from G.
\item train $G$ to minimize $log(1 - D(G(z)))$.
\end{itemize}

\vspace{-2mm}
\begin{equation}
\small
L_{(G,D)} = \min_{D} \min_{G} \left(\mathbb{E}_{x} log D(x) + \mathbb{E}_{z} log (1 - D(G(z)))\right)
\label{eq:loss_gan}
\end{equation}

\begin{figure}[!htb]
\centering
\includegraphics[width=0.3\textwidth]{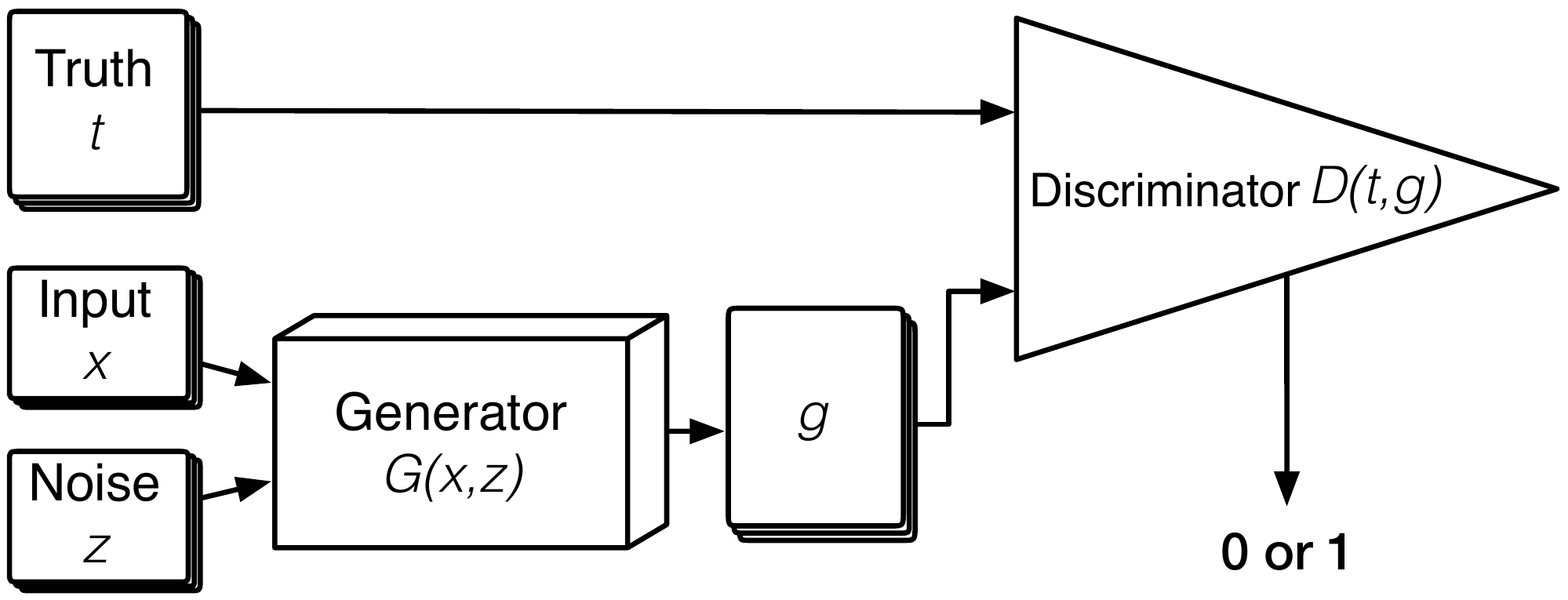}
\caption{Model Overview}
\label{fig:model_overview}
\end{figure}

In contrast to GANs, conditional GANs (also known as cGANs) \cite{mirza2014conditional} learn a mapping by observing both input vector $x$ and random noise vector $z$, denoted as $G(x,z)$, which maps the input $x$ and the noise vector $z$ to $g$, $(x,z) \rightarrow g$ (Figure \ref{fig:model_overview}). The main difference compared to GANs is that the generator and discriminator observe the input vector $x$. Accordingly, the loss function $cL_{(G,D)}$ (Equation \ref{eq:loss_cgan}) and training objectives will be the follows:
\begin{equation}
\small
cL_{(G,D)} = \min_{D} \min_{G} \left(\mathbb{E}_{x,g} log D(x,g) + \mathbb{E}_{x,z} log (1 - D(G(x,z)))\right)
\label{eq:loss_cgan}
\end{equation}

\begin{itemize}
\item train $D$ to maximize the probability of assigning the correct label to both training examples and samples from G.
\item train $G$ to minimize $log(1 - D(G(x,z)))$.
\end{itemize}

In addition, the GAN objective could be further improved with a combined loss function according to \cite{pix2pix2016}, such as adding L1 or L2 distance to the objective, where the discriminator's loss remains unchanged. The objective with L1 distance is
\[
cL_{(G,D)} + \lambda \cdot \mathbb{E}_{x,g,z}[||g - G(x,z)||]
\]

{In an unconditioned GANs, the data is generated without any constrains. With conditional settings, the model is trained with additional information that directly constrains the data generation process, which has been proven to be crucial for image painting and inpainting tasks \cite{zhu2017unpaired}. Conditional settings are particularly important in our context since the input and output images have absolute identical structures.}

\subsection{Features}

In this section, we define the input used for training and inference. The input $x$ includes two parts, i.e., post-placement image $img_{place}$ and connectivity image that represents $Graph(V,E',grids)$ (see Section \ref{sec:pd}). The image $img_{place}$ is generated using the generator implemented based on VPR's interactive mode, where $img_{place} \in \mathbb{R}^{w\times w \times3}$. 

\noindent
\textbf{Color Scheme:} First, a color scheme is used to differentiate the elements in placement and routing. Specifically, the color scheme used in this work is shown in Table \ref{tbl:color_scheme}, which is the default setting used in VPR's interactive mode. Note that other color schemes could be used as well while different elements can be well differentiated using RGB euclidean distance. We show the importance of the color scheme by comparing to using a grayscale image as input in the result section.

\begin{table}[!htb]
\scriptsize
\caption{Color scheme used in post-placement and post-routing images.}
\begin{tabular}{|l|c|c|}
\hline
\multicolumn{1}{|c|}{Color} & $img_{place}$ & $img_{route}$ \\ \hline
White & Routing channels & Out of floor plan \\ \hline
Lightblue & CLB spots & Remaining CLB spots \\ \hline
Pink & Multiplier & Multiplier \\ \hline
Lightyellow & Memory & Memory \\ \hline
Black & Used CLB and IO spots & Used CLB and IO spots \\ \hline
Yellow2purple gradient & - & Routing utilization \\ \hline
\end{tabular}
\label{tbl:color_scheme}
\end{table}

\begin{figure}[!htb]
\centering
\begin{minipage}{0.11\textwidth}
  \centering
\includegraphics[width=1\textwidth]{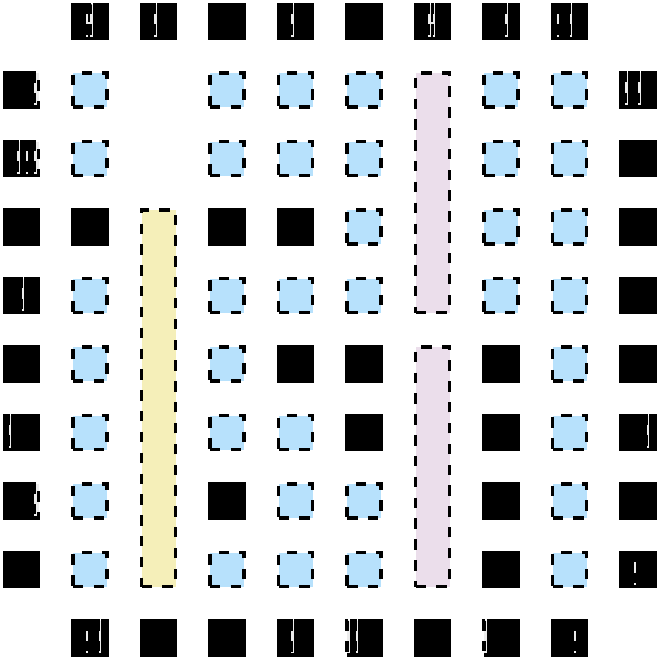}
\end{minipage}
%
\begin{minipage}{0.11\textwidth}
  \centering
\includegraphics[width=1\textwidth]{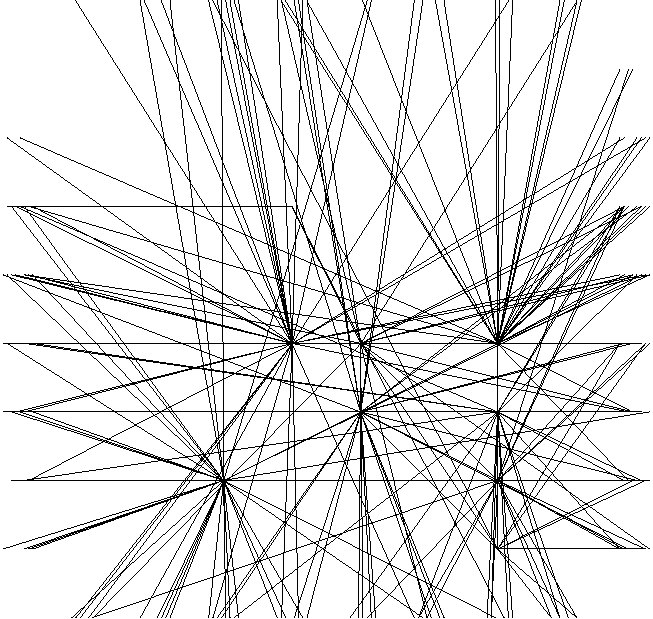}
\end{minipage}
%
\begin{minipage}{0.11\textwidth}
  \centering
\includegraphics[width=1\textwidth]{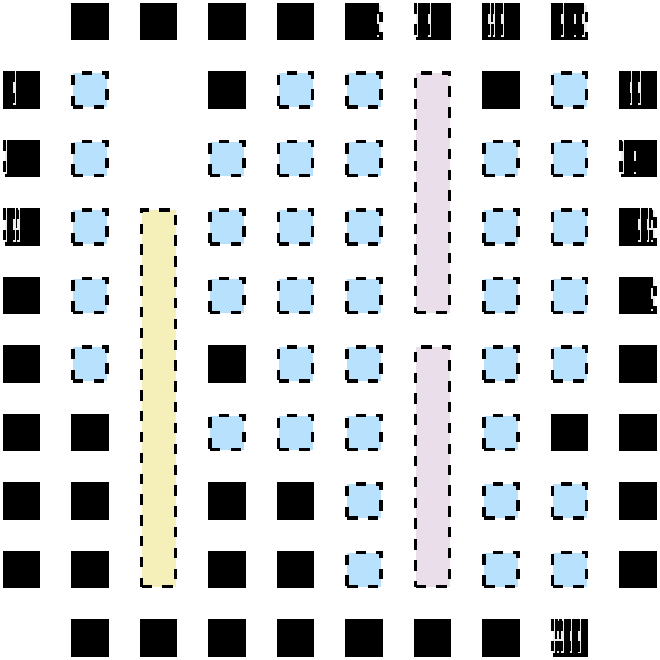}
\end{minipage}
\begin{minipage}{0.11\textwidth}
  \centering
\includegraphics[width=1\textwidth]{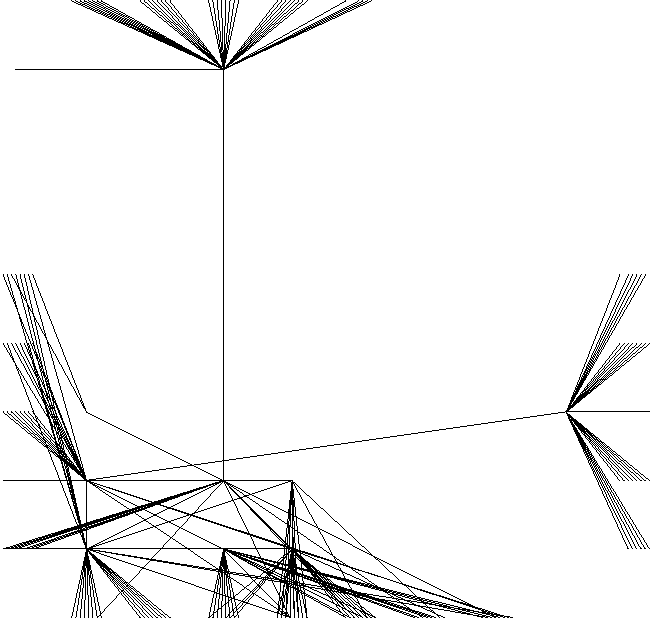}
\end{minipage}
\caption{Connectivity images of based on two different placements results.}
\label{fig:connectivity}
\end{figure}

\noindent
\textbf{Connectivity Image:} In order to use the connectivities of features $Graph(V,E',grids)$ in the neural network, we convert $Graph(V,E',grids)$ into connectivity image, namely $img_{connect}$. Each edge in $E'$ connects two nodes in $V$, which have specific 2-D locations. Drawing edges in $E'$ according to these locations constructs $img_{connect}$. For example, the connectivity images of two different placement results are shown in Figure \ref{fig:connectivity}. Moreover, the connectivity image has the same dimensions as $img_{place}$ but with only one channel, i.e., $\mathbb{R}^{w\times w \times1}$. Note that both $img_{place}$ and $img_{connect}$ are first generated in vector images, and will be converted to bitmap images for training and inference.

\noindent
\textbf{Resolution:} Finally, the dimension $w$ of the input images $img_{place}$ and $img_{connect}$, have to be adjusted based on the size of the floor plan. The goal is to maintain the actual placement structure of $img_{place}$, and differentiates all the elements in the netlist. Specifically, we adjust the resolution of $img_{place}$ such that the dimension of each placement element $\geq$2$\times$2. Note that $img_{place}$ and $img_{connect}$ are vector graphics that can be converted to arbitrary resolution bitmap images. In this work, $w$ is set to be 256.

Hence, the input feature $x$:
\vspace{-2mm}
\[
x = stack(img_{place}, \lambda \cdot img_{connect}), x \in \mathbb{R}^{256\times 256 \times 4}
\]

\vspace{-3mm}

\subsection{Architecture}

\begin{figure}[!htb]
\centering
\includegraphics[width=0.45\textwidth]{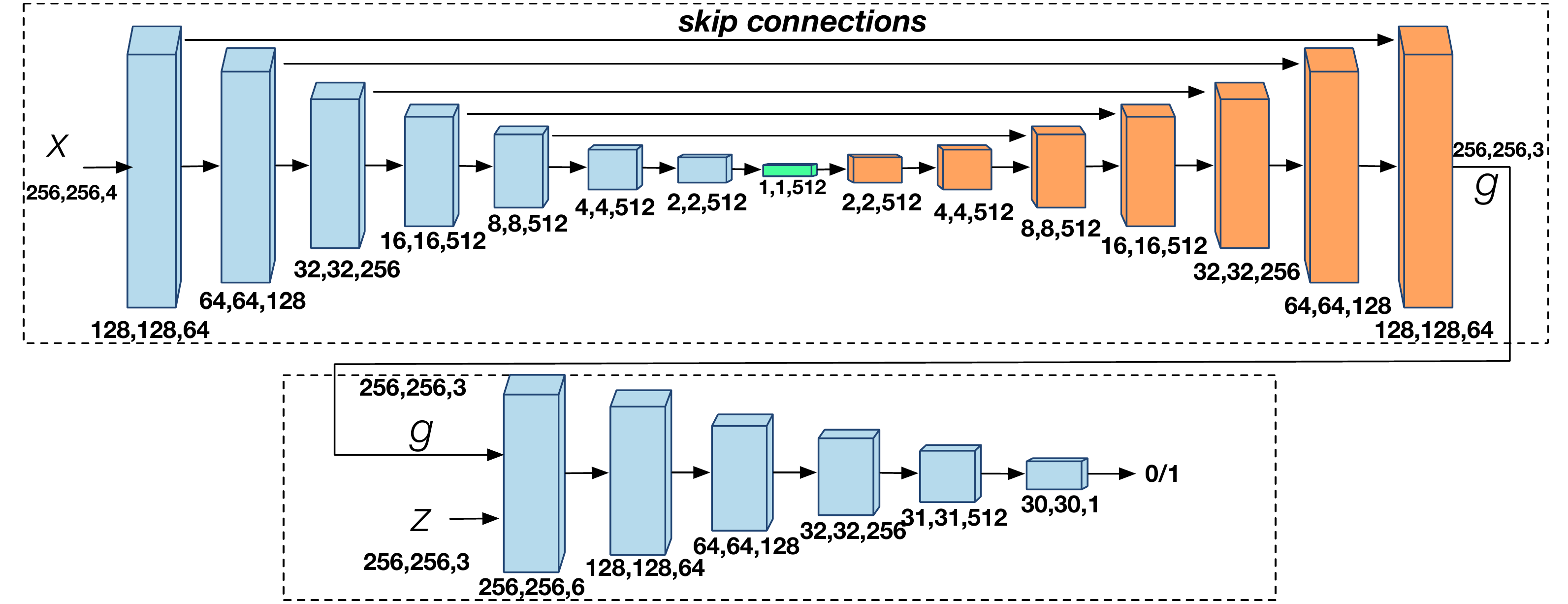}
\caption{Architecture of our conditional GAN model with skip connections.}
\label{fig:network}
\end{figure}

The conditional GANs architecture used in this work is shown in Figure \ref{fig:network}. The generator takes input $x$ and produces output $g$ that includes convolutional and deconvolutional layers only. The discriminator detects whether the output of generator is true or fake, which includes six layers convolutional layers (with batch normlization) followed by sigmoid function for binary classification. 

\noindent
\textbf{Skips in FCNs:} The skip connections in the FCN are shown to important to passing the image structure from the input to the output \cite{pix2pix2016}\cite{xie2018routenet}. The main idea behind is that the network for translating images requires that the information of the input image passes through all the layers. Specifically, in the case of translating $img_{place}$ to $img_{route}$, the input and output share the location of all the image structure edges. The skip connections are shown in Figure \ref{fig:network}.

\subsection{Training}

\begin{figure}[!htb]
\centering
\includegraphics[width=0.4\textwidth]{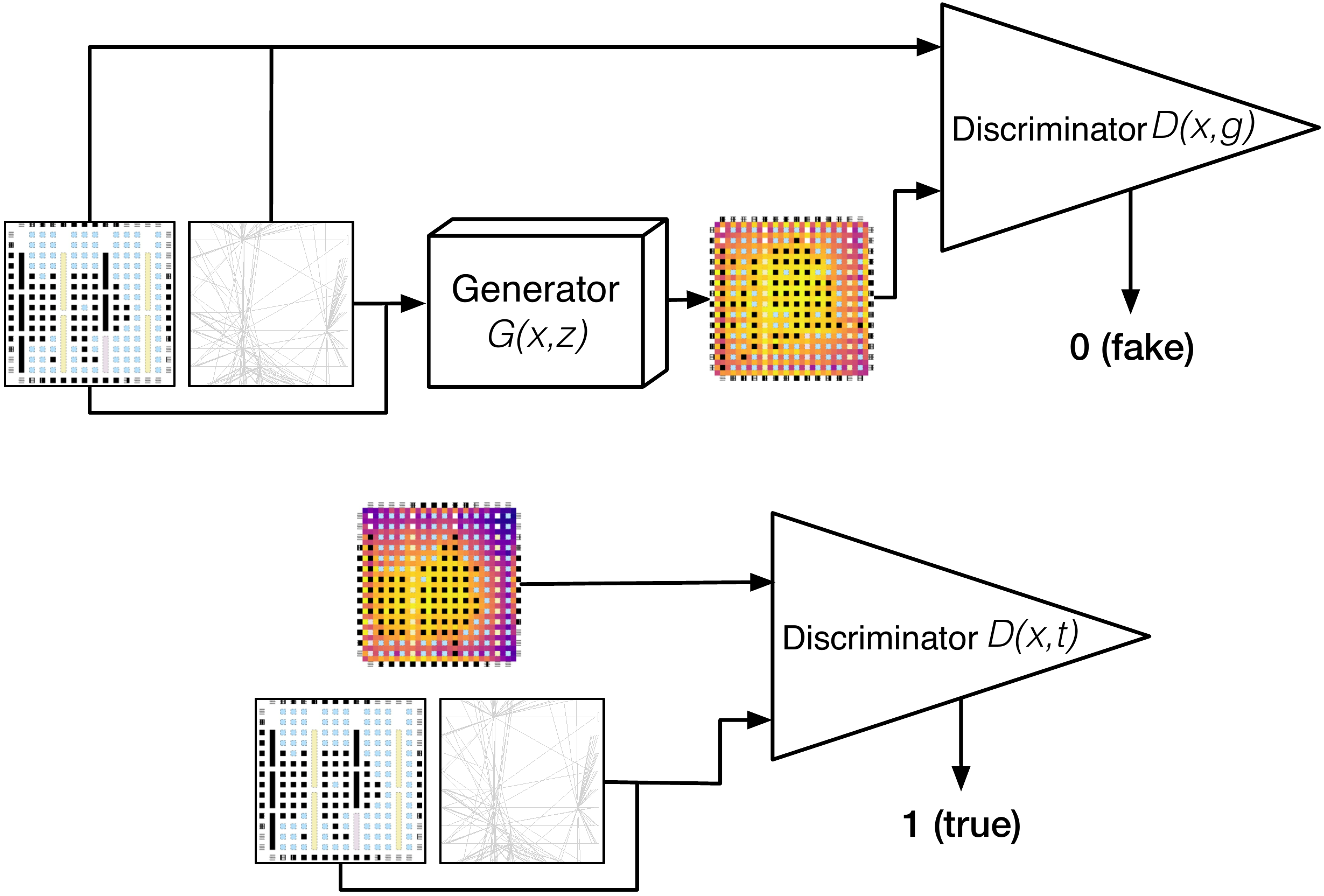}
\caption{Discriminator is trained learns to classify between fake and true combinations ($x,g$) where $x$ is the input and $g$ is the output. Generator is trained to generate images that discriminator cannot distinguish true/fake images.}
\label{fig:training}
\end{figure}

The discriminator is trained with the output images produced by the generator to distinguish the input-truth and input-output pairs. The weights of the discriminator are updated by back-propagation based on the classification error between the input-truth and input-output pairs. The generator is trained by updating its weights based on the difference between input and truth images while the weights of the generator are updated by the output of the discriminator as well (Figure \ref{fig:training}).

\vspace{-3mm}
\section{Results}
 
We evaluate the proposed approach using eight designs listed in Table \ref{tbl:result}, obtained from VTR 8.0 \cite{luu2014vtr}. The image generator is implemented in C++ based on VPR \cite{luu2014vtr}. The input images are first generated as vector graphics and are converted to JPEG with $w$=256. The training and inference of cGAN are implemented in Python3 using Tensorflow. The experimental results are obtained using a machine with a 10-core Intel Xeon operating at 2.5 GHz, 1 TB RAM, and one Nvidia 1080Ti GPU. The learning rate is 0.0002 using Adam optimizer, where the momentum term $\beta_1$=0.5 and $\beta_2$=0.999 with $\epsilon$=$10^{-8}$. The L1 weight is 50 and $\lambda$ is set to 0.1. The number of training epochs is 250 with bath size 1. The training time is 2-3 hours and inference takes about 0.09 second per image. 

\noindent
{\bf Datasets:} The placement results are generated by sweeping the VPR placement options, including \texttt{seed, ALPHA\_T, INNER\_NUM} and \texttt{place\_algorithm}. The ground truth images are collected with these placement options with default VPR settings.

\subsection{Quality of Routing Forecasts}
 
Our dataset includes 1500 input-output image pairs. The input images include post-placement image $img_{place}$ and connectivity image $img_{connect}$. The ground truth images are $img_{route}$ generated after VPR default routing. The speedup is measured using the magnitude of routing runtime divided by inference time since the routing runtime varies based on different placement. Two accuracy metrics are used to evaluate our approach. First, \textit{per-pixel} accuracy between the generated image and ground truth image is used to evaluate the generated image quality (Acc.$_{1}$ and Acc.$_{2}$ in Table \ref{tbl:result}). Second, \textit{Top10} indicates the top-10 accuracy for finding min-congestion placements within the testing set. For example, Top10=80\% means that there are eight placements are truly top 10 among the ten selected ones. 

The quality of the routing forecasts is evaluated using eight designs, shown in Table \ref{tbl:result}. Two training strategies are applied in this work. \textbf{1)} The training set includes all the images \underline{except} the testing design. This makes sure that the training dataset has no overlap with the testing dataset. In other words, this applies inference on unseen designs. The accuracy is shown in Acc.$_{1}$. 2) To further improve the robustness of our approach, we update the model trained with the first training strategy using only \textbf{ten} input-output image pairs from the testing design, which takes the advantages of transfer learning. The testing accuracy improved, particularly for the SHA design. One observation is that forecasting for the smallest designs (i.e., diffeq1 and diffeq2) is less accurate than the larger designs. The reason could be that the placement and routing algorithms can find the near-optimal solution(s) with most tool options for small designs, which makes the dataset very unbalanced. Top10 results in Table \ref{tbl:result} are obtained using the second strategy.

\begin{table}[t]
\scriptsize
\caption{Experimental results obtained using eight designs. {Acc.$_{1}$ and Acc.$_{2}$ are per-pixel accuracy obtained using two training strategies. \#P(\# placements) indicates the number of input and output image pairs.}}
\begin{tabular}{|l|l|l|l|l|l|l|l|}
\hline
\textbf{Design} & \multicolumn{1}{c|}{\bf \#LUTs} & \multicolumn{1}{c|}{\bf \#FF} & \multicolumn{1}{c|}{\bf \#Nets} & \multicolumn{1}{c|}{\bf \# P} & \multicolumn{1}{c|}{\bf Acc.$_{1}$} & \multicolumn{1}{c|}{\bf Acc.$_{2}$} & \multicolumn{1}{c|}{\bf Top10}\\ \hline
\textit{diffeq1} & 563 & 193 & 2,059 & 200 & 67.2\% & 68.9\% & 50\% \\ \hline
\textit{diffeq2} & 419 & 96 & 1,560 & 200 & 65.3\% & 65.9\%  & 40\%\\ \hline
\textit{raygentop} & 1,920 & 1,047 & 5,023 & 200 & 68.1\% & 77.1\%  & 70\% \\ \hline
\textit{SHA} & 2,501 & 911 & 10,910 & 200 & 43.3\% & 61.0\%  & 40\%\\ \hline
\textit{OR1200} & 2,823 & 670 & 12,336  & 200 & 64.6\% & 67.6\%  &  {90\%}\\ \hline
\textit{ode} & 5,488 & 1,316 & 20,981 & 200 & 74.9\% & 75.9\% &  {80}\%\\ \hline
\textit{dcsg} & 9,088 & 1,618 & 36,912 & 200 & 71.4\% & 85.4\%  & {80}\% \\ \hline
\textit{bfly} & 9,503 & 1,748 & 38,582 & 200 &  71.5\%& 76.5 \%&  70\%\\ \hline
\end{tabular}
\vspace{-3mm}
\label{tbl:result}
\end{table}

\vspace{-3mm}
\subsection{Color Scheme vs. Grayscale}

One of the key input of our cGAN model, $img_{place}$, is an RGB image. While $img_{place}$ is generated, a specific color scheme is used to differentiate the elements for placement. To evaluate the importance of the color scheme, we compare the performance of RGB $img_{place}$ with its grayscale version. The images are converted to grayscale using \texttt{tf.image.rgb\_to\_grayscale}\footnote{\url{ https://www.tensorflow.org/api_docs/python/tf/image/rgb_to_grayscale}}. The average per-pixel accuracy drops 3-5\%, and the inference images are mostly "brighter" than the ground truth images. This makes it less accurate for the inputs that their outputs are less congested. This also saves $\sim$20\% training time and $\sim$50\% for inference. While the training and inference runtime is not critical in this context, we always choose colored placement image as inputs.

\subsection{Analysis of L1 and skip connections}\label{sec:l1_skip}

\begin{figure}[!htb]
\centering
\begin{minipage}{0.17\textwidth}
  \centering
\includegraphics[width=1\textwidth]{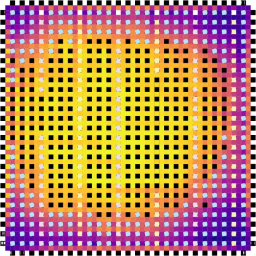}
\subcaption{Truth}\label{fig:skip_L1_1}
\end{minipage}
\hspace{1mm}
\begin{minipage}{0.17\textwidth}
  \centering
\includegraphics[width=1\textwidth]{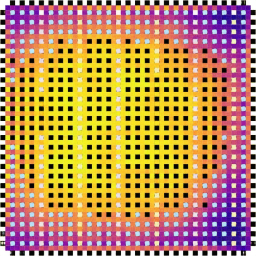}
\subcaption{L1+all skip}\label{fig:skip_L1_2}
\end{minipage}
\hspace{1mm}
\begin{minipage}{0.17\textwidth}
  \centering
\includegraphics[width=1\textwidth]{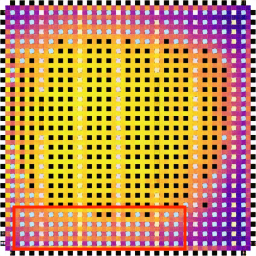}
\subcaption{w/o L1+all skip}\label{fig:skip_L1_3}
\end{minipage}
\hspace{1mm}
\begin{minipage}{0.17\textwidth}
  \centering
\includegraphics[width=1\textwidth]{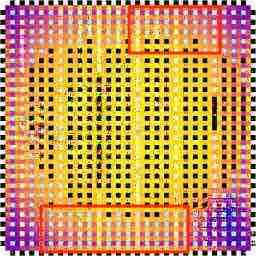}
\subcaption{L1+Single skip}\label{fig:skip_L1_4}
\end{minipage}
\caption{Comparing the ground truth image with generated images using three different models using OR1200.}
\vspace{-3mm}
\label{fig:skip_L1}
\end{figure}

\begin{figure}[!htb]
\centering
\begin{minipage}{0.223\textwidth}
  \centering
    \includegraphics[width=1\textwidth]{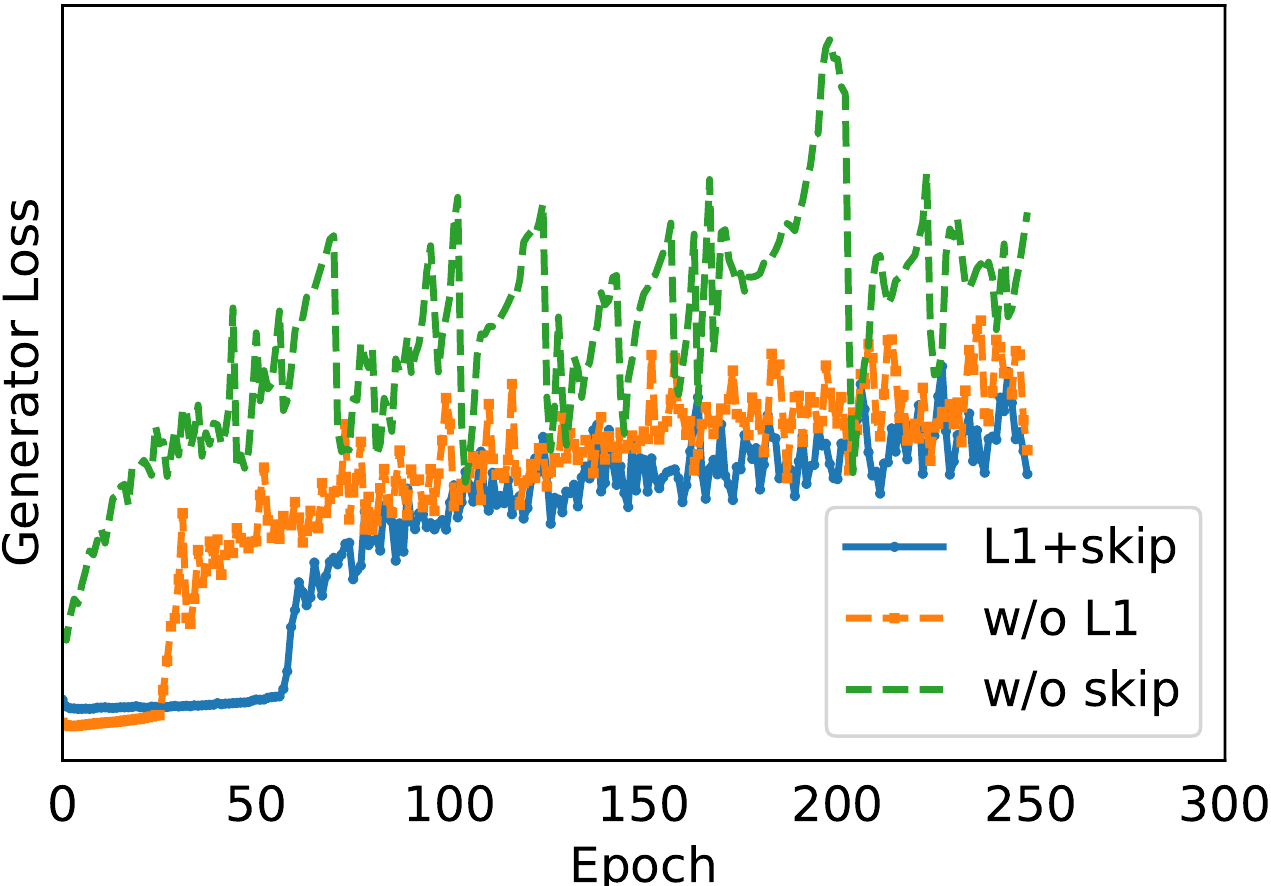}
\subcaption{Generator training loss.}
\end{minipage}
\begin{minipage}{0.223\textwidth}
 \centering
\includegraphics[width=1\textwidth]{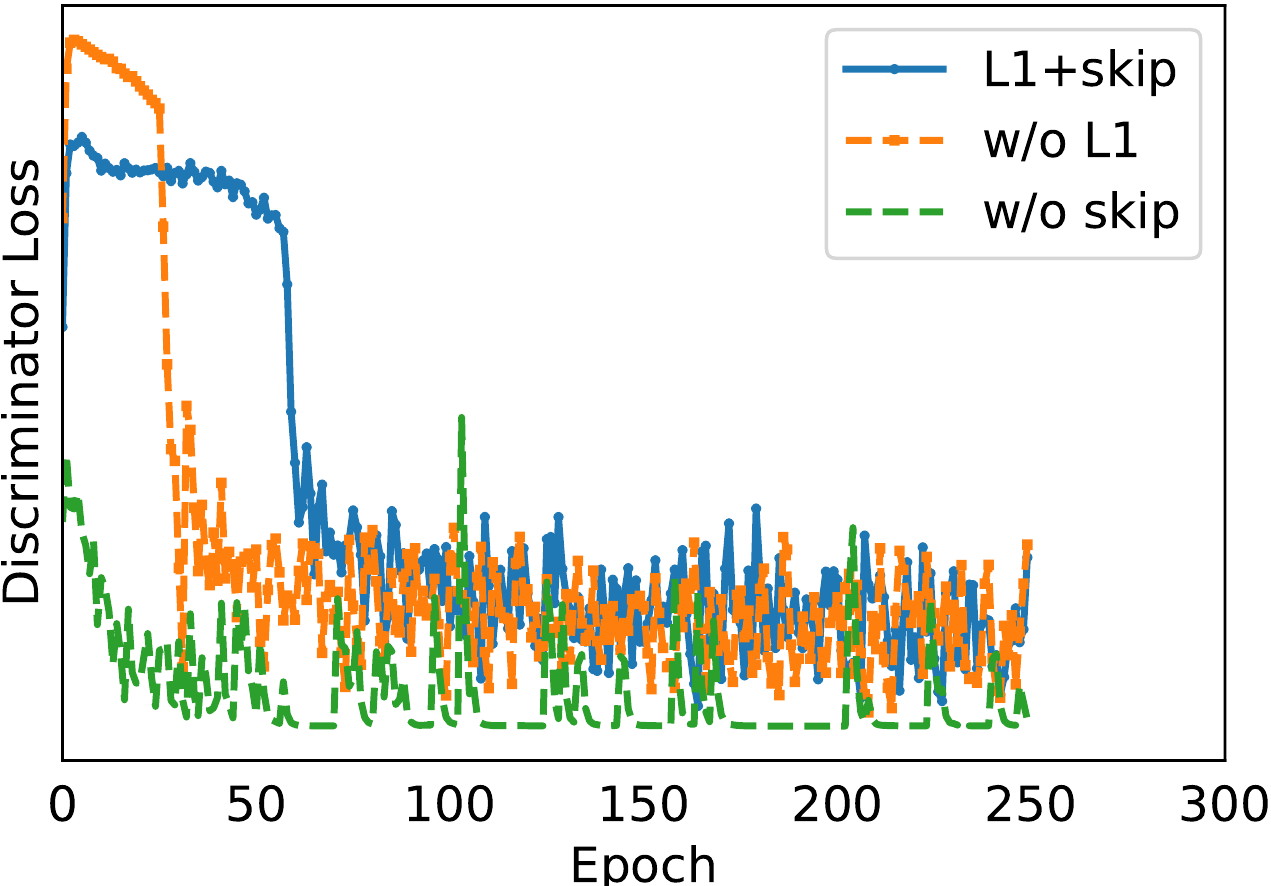}
\subcaption{Discriminator training loss.}
\end{minipage}
\caption{Evaluating the effects of L1 and skip connections by comparing a) generator training loss and b) discriminator training loss.}
\label{fig:skip_connection_loss}
\vspace{-3mm}
\end{figure}

We analyze the effectiveness of using L1 in the loss function and the skip connections in the generator using OR1200 design. First, we compare the inference results by forecasting routing utilization of one placement, shown in Figure \ref{fig:skip_L1}. The ground truth image and the inference image with full skip connections and L1 are shown in Figures \ref{fig:skip_L1_1} and \ref{fig:skip_L1_2}, where two images are almost identical. Using the same architecture but without L1 for training, a mispredicted region is clearly found in Figure \ref{fig:skip_L1_2}. Xie et al. demonstrated that using a single skip connection in the FCN is sufficient for hotspot prediction \cite{xie2018routenet}. However, we observe that it is necessary to connect all the convolutional and deconvolutional layers (see Figure \ref{fig:network}) for forecasting the entire routing heat map. As shown in Figure \ref{fig:skip_L1_4}, we can clearly see the mispredicted regions and a large number of noises over the inference image. We further analyze the effects of L1 and skip connections by measuring the training loss of generator and discriminator. The results are included in Figure \ref{fig:skip_connection_loss}. We observe that loss functions are optimized smoothly if both L1 and skip connections are used, and the training losses are aggressively optimized with relative large noises. These mostly lead to over- or under-fitting problem. In addition, there are more training noise if the model has a single skip connection compared to without L1. This explains why the model without skip connections generates worse routing heat map compared to without L1; and why L1+skip generates the best results among these three options.

\begin{figure*}[!htb]
Place$~~~$
\centering
\begin{minipage}{0.12\textwidth}
  \centering
\includegraphics[width=1\textwidth]{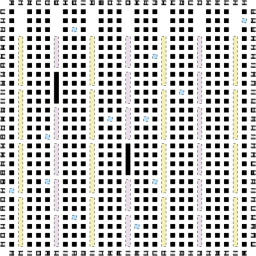}
\end{minipage}
\hspace{1mm}
\begin{minipage}{0.12\textwidth}
  \centering
\includegraphics[width=1\textwidth]{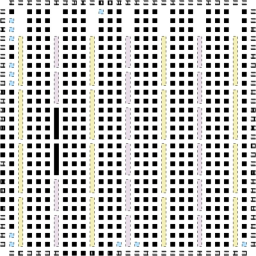}
\end{minipage}
\hspace{1mm}
\begin{minipage}{0.12\textwidth}
  \centering
\includegraphics[width=1\textwidth]{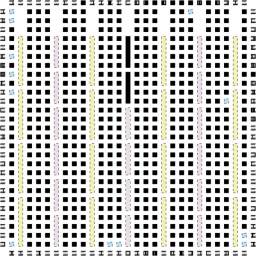}
\end{minipage}
\hspace{1mm}
\begin{minipage}{0.12\textwidth}
  \centering
\includegraphics[width=1\textwidth]{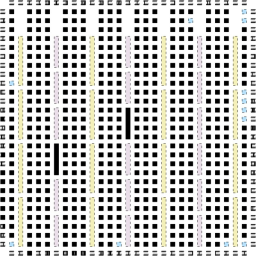}
\end{minipage}
\hspace{1mm}
\begin{minipage}{0.12\textwidth}
  \centering
\includegraphics[width=1\textwidth]{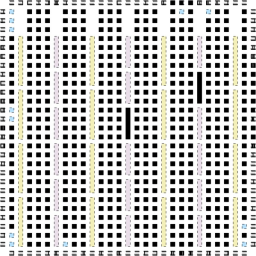}
\end{minipage}
\\
Output
\begin{minipage}{0.12\textwidth}
  \centering
\includegraphics[width=1\textwidth]{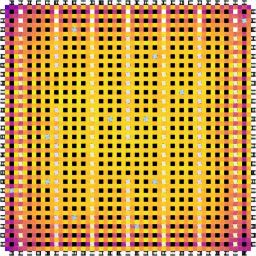}
\end{minipage}
\hspace{1mm}
\begin{minipage}{0.12\textwidth}
  \centering
\includegraphics[width=1\textwidth]{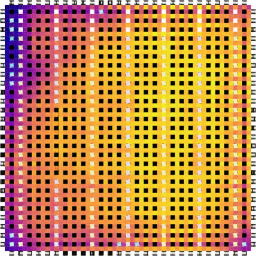}
\end{minipage}
\hspace{1mm}
\begin{minipage}{0.12\textwidth}
  \centering
\includegraphics[width=1\textwidth]{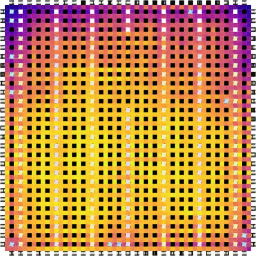}
\end{minipage}
\hspace{1mm}
\begin{minipage}{0.12\textwidth}
  \centering
\includegraphics[width=1\textwidth]{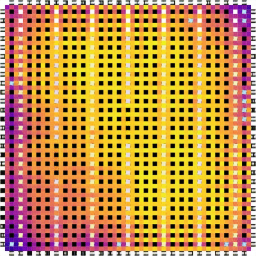}
\end{minipage}
\hspace{1mm}
\begin{minipage}{0.12\textwidth}
  \centering
\includegraphics[width=1\textwidth]{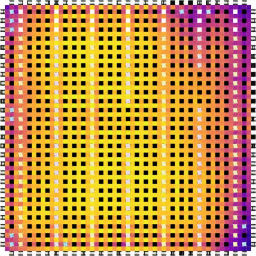}
\end{minipage}
\\
Truth$~$
\begin{minipage}{0.12\textwidth}
  \centering
\includegraphics[width=1\textwidth]{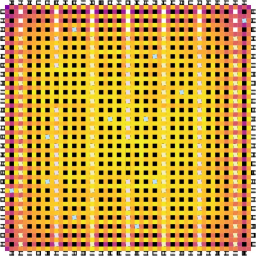}
\subcaption{Overall-max}\label{fig:Truth1}
\end{minipage}
\hspace{1mm}
\begin{minipage}{0.12\textwidth}
  \centering
\includegraphics[width=1\textwidth]{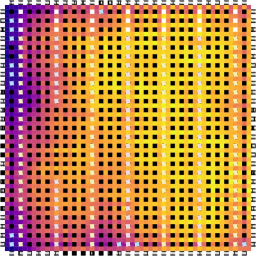}
\subcaption{Overall-min}\label{fig:Truth2}
\end{minipage}
\hspace{1mm}
\begin{minipage}{0.12\textwidth}
  \centering
\includegraphics[width=1\textwidth]{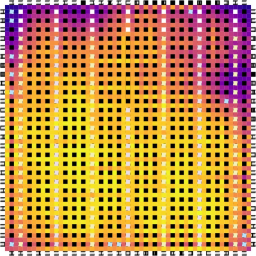}
\subcaption{Upper-min}\label{fig:Truth3}
\end{minipage}
\hspace{1mm}
\begin{minipage}{0.12\textwidth}
  \centering
\includegraphics[width=1\textwidth]{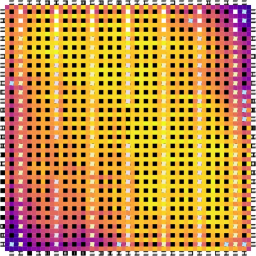}
\subcaption{Lower-min}\label{fig:Truth4}
\end{minipage}
\hspace{1mm}
\begin{minipage}{0.12\textwidth}
  \centering
\includegraphics[width=1\textwidth]{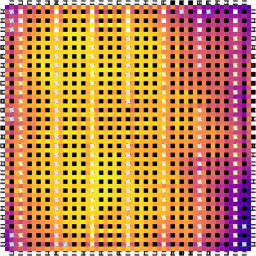}
\subcaption{Right-min}\label{fig:Truth5}
\end{minipage}
\caption{Constrained placement exploration by inference using ODE -- Obtained placement solutions with objectives a) overall max-congestion, b) overall min-congestion, c) min-congestion at the upper side, d) min-congestion at the lower side, and e) min-congestion at the right-hand side of the floor plan.}
\vspace{-4mm}
\label{fig:place_enum}
\end{figure*}

\subsection{Applications}\label{sec:app}

While in Table \ref{tbl:result} column \textit{Top10}, it is demonstrated that the proposed approach can effectively explore the placement solutions and find the placements with lowest routing congestion. To further demonstrate the advantages of fully forecasting routing heat map, the proposed approach is leveraged to solve the following problems:

\noindent
\textbf{Constrained placement exploration}: The goal is to search for placement solutions in the dataset of \textit{ode} design that have the highest congestion, lowest congestion, high congested at the top, bottom, and right regions of the floor plan, shown from left to right in Figure \ref{fig:place_enum}, respectively. Moreover, the routing density of less congested regions also well correct to the ground truth. This demonstrates that our approach can accurately predict the routing density of all the channels.

\noindent
\textbf{Visualizing the simulated annealing placement algorithm}: The proposed approach is applied to visualize the routing utilization \textit{on-the-fly} during placement. This allows us to visualize how the density of routing channels are changed while the design is ''being placed''. Here, we apply to the classic simulation annealing based placement algorithm implemented in VPR. The real-time forecast results (GIF videos) are included\footnote{\url{https://ycunxi.github.io/cunxiyu/dac19_demo.html}}.

\section{Acknowledgments}
This work is funded by Intel Corporation under the ISRA Program. The authors would like to thank Dr. Wang Zhou and Dr. Gi-Joon Nam at IBM Thomas J. Watson Research Center for the invaluable discussions.

\small
\bibliographystyle{IEEEtran}
\bibliography{synthesis}

\end{document}